\documentclass[conference,a4paper]{IEEEtran}
\usepackage{graphicx,color}
\usepackage{cite}
\usepackage{amssymb,amsmath,mathtools}
\usepackage{amsfonts}

\usepackage{bm,bbm}
\usepackage{booktabs}
\usepackage[caption=false,font=footnotesize]{subfig}
\graphicspath{{./figs/}}
\usepackage{algorithm}
\usepackage{algpseudocode}

\newcommand{\Hseq}{\left\{\bm{H}[k]\right\}_{k}}
\newcommand{\Vseq}{\left\{\bm{V}[k]\right\}_{k}}

\begin{document}
\title{Frame-Capture-Based CSI Recomposition \\ Pertaining to Firmware-Agnostic WiFi Sensing}
\author{
	\IEEEauthorblockN{
		\normalsize Ryosuke Hanahara\IEEEauthorrefmark{1},
		\normalsize Sohei Itahara\IEEEauthorrefmark{1},
		\normalsize Kota Yamashita\IEEEauthorrefmark{1},
		\normalsize Yusuke Koda\IEEEauthorrefmark{2} \\
		\normalsize Akihito Taya\IEEEauthorrefmark{1}\IEEEauthorrefmark{3},
		\normalsize Takayuki Nishio\IEEEauthorrefmark{1}\IEEEauthorrefmark{4}, and
		\normalsize Koji Yamamoto\IEEEauthorrefmark{1}
	}\\
	\IEEEauthorblockA{
		\IEEEauthorrefmark{1}\small Graduate School of Informatics, Kyoto University,
		Yoshida-honmachi, Sakyo-ku, Kyoto 606-8501, Japan \\
		\IEEEauthorrefmark{2}\small Centre for Wireless Communications, University of Oulu, 90014 Oulu, Finland\\
		\IEEEauthorrefmark{3}\small College of Science and Engineering, Aoyama Gakuin University, 5-10-1 Fuchinobe, Chuo-ku, Sagamihara-shi,Kanagawa 252-5258, Japan\\
		\IEEEauthorrefmark{4}\small School of Engineering, Tokyo Institute of Technology, Ookayama, Meguro-ku, Tokyo, 152-8550, Japan \\ 
		(e-mail: kyamamot@i.kyoto-u.ac.jp) \\
	}
}

\maketitle
\begin{abstract}
	With regard to the implementation of WiFi sensing agnostic according to the availability of channel state information (CSI), we investigate the possibility of estimating a CSI matrix based on its compressed version, which is known as beamforming feedback matrix (BFM). Being different from the CSI matrix that is processed and discarded in physical layer components, the BFM can be captured using a medium-access-layer frame-capturing technique because this is exchanged among an access point (AP) and stations (STAs) over the air.	This indicates that WiFi sensing that leverages the BFM matrix is more practical to implement using the pre-installed APs. However, the ability of BFM-based sensing has been evaluated in a few tasks, and more general insights into its performance should be provided. To fill this gap, we propose a CSI estimation method based on BFM, approximating the estimation function with a machine learning model. In addition, to improve the estimation accuracy, we leverage the inter-subcarrier dependency using the BFMs at multiple subcarriers in orthogonal frequency division multiplexing transmissions. Our simulation evaluation reveals that the estimated CSI matches the ground-truth amplitude. Moreover, compared to CSI estimation at each individual subcarrier, the effect of the BFMs at multiple subcarriers on the CSI estimation accuracy is validated.
\end{abstract}

\section{Introduction}
Wireless local area networks (WLANs) rapidly increase popularity with the improvement of their throughput.
One of the key technologies of WLANs to realize high throughput is multiple-input multiple-output (MIMO).
Along with orthogonal frequency division multiplexing (OFDM), MIMO provides channel state information (CSI) for each antenna pair at each subcarrier.
In recent times, CSIs provided by WLAN systems have attracted interest for MIMO communications and WiFi sensing due to their fine-granularity for characterizing the surrounding environment.
As WiFi sensing can be performed without any additional equipment rather than WLAN systems, it can be deployed at low cost without considering additional sensing devices, e.g., radars and surveillance cameras.

Despite their low cost, CSI-based WiFi sensing methods require practitioners to extract CSI matrices from wireless chipsets because the CSI matrix is processed and discarded in physical (PHY) layer components.
This procedure needs additional firmware implementation, and such firmware performs on a few wireless chipsets and protocols \cite{gringoli2019free,halperin2011tool}, which limit the applicability of the CSI-based WiFi sensing to few WiFi devices.

\begin{figure}[t]
	\centering
	\includegraphics[width=1.0\columnwidth]{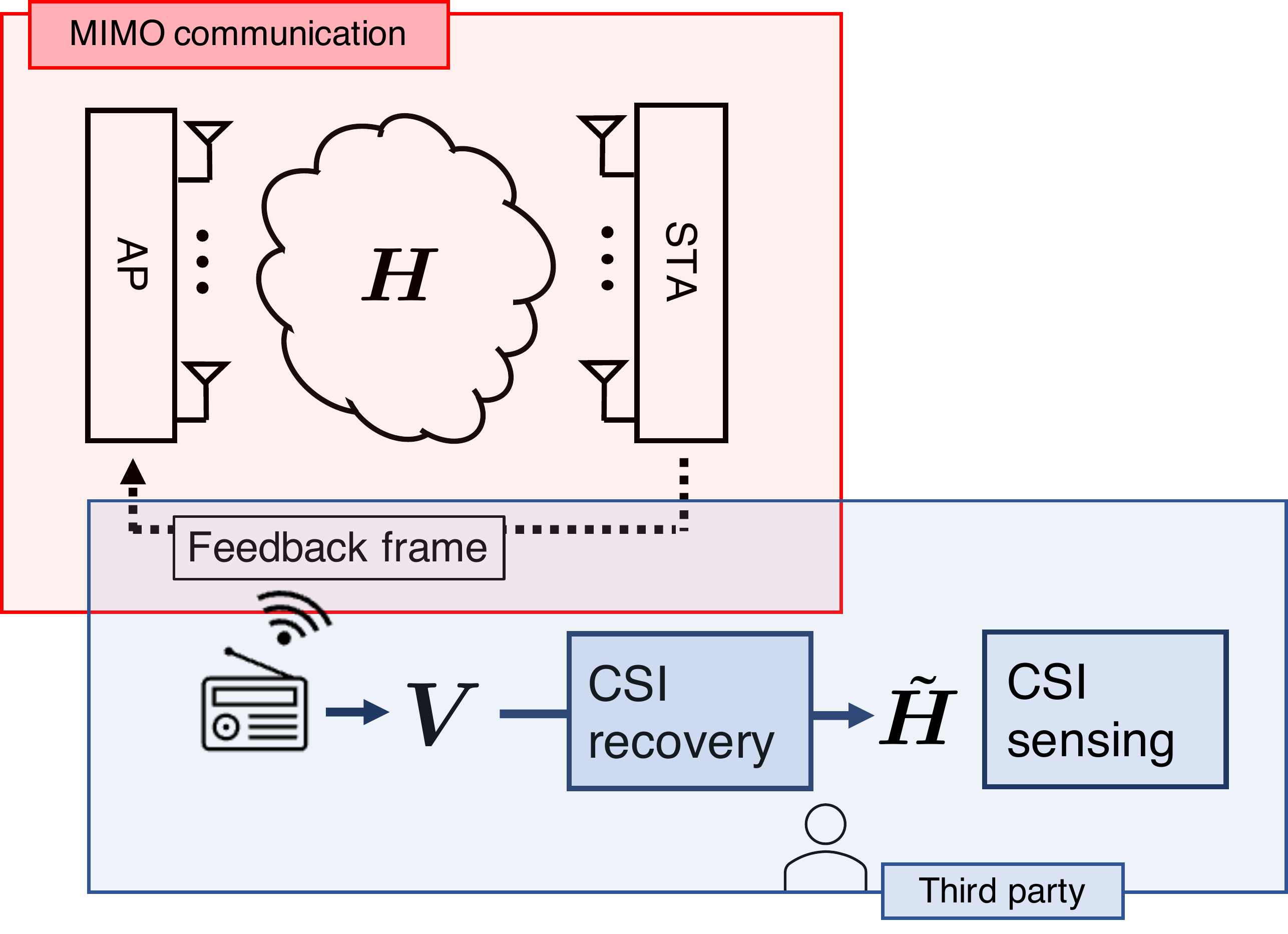}
	\caption{System model of BFM-based sensing.
		To obtain BFM information, the third party captures BFM packets, which are transported from the receiver to the transmitter for efficient MIMO communication.
		Thus, the third party that performs BFM-sensing does not need to access the transmitter or receiver.
	}
	\label{fig:system_model}
\end{figure}

To alleviate this restriction, few studies \cite{miyazaki2019initial, takahashi2019dnn} performed WiFi sensing without explicitly extracting CSI from PHY layer components, but rather leveraged a beamforming feedback matrix (BFM) \cite{11ac}, which is referred to as a compressed version of CSI.
However, WiFi access points (APs) and stations (STAs) are mandated to exchange the BFM without encryption to perform OFDM-MIMO transmissions using the IEEE 802.11ac/ax standard \cite{11ac, 11ax}.
This indicates that the BFM is extracted from the PHY-layer components by default and is stored in a corresponding field in medium access (MAC)-layer frames exchanged among APs or STAs.
This enables practitioners to capture BFMs using MAC frame capturing tools (Fig.~\ref{fig:system_model}), where the third-party user can carry out WiFi sensing regardless of the lack of access to the PHY layer components of the transmitter and receiver.

However, existing BFM-based WiFi sensing studies \cite{miyazaki2019initial, takahashi2019dnn} have addressed only a few sensing tasks, while CSI-based WiFi sensing has addressed several tasks \cite{ma2019wifi, wang2019survey}.
Specifically, studies related to BFM-based WiFi sensing are limited to human detection and localization\cite{miyazaki2019initial, takahashi2019dnn}, whereas the CSI-based WiFi sensing acquires various usage models, e.g., human activity recognition \cite{chen2018passive,wang2016wi-hear,ali2017recognizing}, device localization \cite{kotaru2015spotfi}, and human vital sensing \cite{zeng2019farsense}.
Hence, there is a knowledge gap between BFM-based sensing and CSI-based WiFi sensing considering that we are not aware of the feasibility of the replacement of the CSI matrix with the BFM to perform various sensing tasks.
This leads us to a research question of how much BFMs are informative to replace CSI matrices to carry out sensing tasks?

To fill this gap and provide a general insight into this question, we explore the possibility of estimating the CSI from BFM and evaluate its accuracy.
The intuition behind this is that, if the estimation is possible, the replacement of the CSI with the BFM can also carry out various WiFi sensing tasks.
To this end, we propose a machine learning (ML)-based CSI estimation from BFM and evaluate the estimation accuracy.
In the proposed scheme, we estimate a function from BFM to CSI by approximating the function with the ML model.

To improve the estimation accuracy, we leverage an inter-subcarrier dependency and estimate the function that outputs CSI of a single subcarrier from the BFMs at multiple subcarriers (hereinafter referred to as \textit{the subcarrier-integrated estimation}).
Further, as an input layer of the ML model, we use ``frequency-directional convolutional layer,''
that outputs the convolution result of the BFMs and a kernel filter in direction of frequency.
The simulation evaluation reveals that the estimated frequency-selective CSI matches the ground-truth amplitude
compared with CSI estimation based on a BFM for each subcarrier (hereinafter referred to as \textit{the subcarrier-individual estimation}).
In this study, as a firsthand approach, the amplitude of the CSI elements is estimated.
Although we limit the discussion about the amplitude elements, this contributes to answering the aforementioned questions because several sensing tasks \cite{wang2017carm, li2016csi, chen2018passive} were conducted by the amplitude of CSI.

Another by-product benefit to investigate the CSI estimation is to debunk the potential privacy leakage from exchanging BFMs.
As the CSI-based WiFi sensing carries out various tasks, as discussed above, CSI will be considered as private sensitive information \cite{cominelli2021ieee}.
Accordingly, from the privacy perspective, it will be questionable whether the BFMs should be exchanged over the air such that the third-party user can capture them using a frame-capturing tool (Fig.~\ref{fig:system_model}).
As an intuitive example, consider that the AP and STA are public or held by parties other than the third-party user.
The third-party user may succeed in estimating the trajectory of another person of interest by leveraging the existing CSI-based WiFi sensing methods while alternatively using the BFMs as an input.
Hence, we believe that to shed light on the potential privacy risk from exchanging BFMs, the above question (i.e., how much BFMs are informative to replace CSI matrices to carry out sensing tasks?) should be addressed.

The contribution of this study is as follows:
We explore the possibility of the CSI estimation based on its compressed version (i.e., BFM) to gain insight into the capability of the BFM to replace the original CSI for various sensing tasks.
To provide a concrete design for estimation, we have used a ML technique where the ML model accepts a BFM and estimates the original CSI.
Our numerical evaluation reveals that the estimated amplitude element in the CSI matrix well matches the ground-truth amplitude, which shows the feasibility of the aforementioned CSI estimation.
To the best of our knowledge, this insight has not been provided in previous studies with regard to WiFi sensing.

The rest of the study is organized as follows:
Section \ref{sec:SystemModel} describes the CSI in a MIMO environment and a system for feedback.
Section \ref{sec:csi_recovery} explains ML-based methods to recover CSI from BFMs. 
Section \ref{sec:ml_algorithms} shows the evaluation results of estimation accuracy. 
Section \ref{sec:conclusion} concludes the study.

\section{Preliminaries of OFDM-MIMO}
\label{sec:SystemModel}
We introduce an OFDM-MIMO communication system using beamforming techniques based on explicit feedback.
The system model comprises a transmitter and a receiver, where the transmitter sends packets to the receiver via MIMO transmissions being compliant to the IEEE 802.11ac standard.
The receiver measures CSI, computes the BFM from CSI, and sends back the BFM to the transmitter.
In the transmitter, the BFM is used for transmit precoding.

Given the number of antennas of the transmitter and receiver are $N_{\mathrm{t}}$ and $N_{\mathrm{r}}$, respectively,
let us denote the CSI matrix at subcarrier $k$ between the transmitter and receiver to be $\bm{H}[k]$, where $\bm{H}[k] \in \mathbb{C}^{N_{\mathrm{r}}\times N_{\mathrm{t}}}$.
The CSI matrix is estimated for each OFDM subcarrier using a pilot signal (e.g., null data packet in 802.11ac/ax).
Each element of the CSI matrix $h_{ij}[k]$ is represented by its amplitude $a_{ij}[k]$ and phase $\theta_{ij}[k]$ as
\begin{align}
	h_{ij}[k] = a_{ij}[k] \mathrm{e}^{\mathrm{j}\theta_{ij}[k]},
\end{align}
where $i$ and $j$ indicate the indices of the receive and transmit antennas, respectively.
Given the transmitted symbols at $k$th subcarrier $\bm{x}[k]\in \mathbb{C}^{N_{\mathrm{t}}\times 1}$,
the received signal $\bm{y}[k]$ is represented as follows:
\begin{align}
	\bm{y}[k] = \bm{H}[k]\,\bm{x}[k] + \bm{z}[k],
\end{align}
where $\bm{z}[k]\in \mathbb{C}^{N_{\mathrm{r}}\times 1}$ denotes the additive white Gaussian noise.
In the following, $[k]$ is omitted unless otherwise noted.

The 802.11ac/ax standard adopts eigenbeam space division multiplexing (E-SDM) \cite{miyashita2002high} to obtain independent channels from the MIMO system.
In E-SDM, the right singular matrix of the CSI matrix is used for transmit precoding, and the transpose of the left singular matrix is used for decoding at the receiver.
Considering $\bm{V}$ and $\bm{U}^{\mathrm{H}}$ as right singular and left singular matrices, respectively, they satisfy
\begin{align}
	\label{eq:SVD}
	\bm{H} = \bm{U} \bm{\varSigma} \bm{V}^{\mathrm{H}},
\end{align}
where $\bm{U}$ and $\bm{V}$ are unitary matrices and $\bm{\varSigma}$ is a diagonal matrix.
The decomposition of $\bm{H}$ in \eqref{eq:SVD} is known as singular value decomposition (SVD).
In the transmission precoding process, the transmit signal $\bm{x}$ is multiplied by $\bm{V}$, and then $\bm{V}\bm{x}$ is transmitted.
The transmission signal is attenuated in the channel, and $\bm{HVx}$ is received.
The received signal is multiplied by $\bm{U}^\mathrm{H}$; then, we obtain $\bm{U}^\mathrm{H}\bm{HVx}$ as a shaped signal.
With regard to \eqref{eq:SVD}, the shaped signal $\bm{r}$ is represented as
\begin{align}
	\bm{r} = \bm{U}^\mathrm{H}\bm{HVx} = \bm{U}^\mathrm{H}\bm{U}\bm{\varSigma}\bm{V}^\mathrm{H}\bm{Vx} = \bm{\varSigma}\bm{x}.
\end{align}
Considering that $\bm{\varSigma}$ is a diagonal matrix, the E-SDM provides multiple independent channels from the MIMO system.

As explained previously, in the E-SDM, for transmit precoding, the transmitter requires a right singular matrix of the CSI matrix, which is referred to as BFM.
To inform the BFM, the receiver applies SVD to the CSI matrix and obtains the BFM.
The BFM is calculated for each OFDM subcarrier using the corresponding CSI matrix of the subcarrier.
Subsequently, the receiver transmits BFM frames, which contain the BFM, to the transmitter.
Fortunately, the BFM packets are transmitted without encryption; therefore, one can obtain BFM using arbitrary WLAN devices by capturing the BFM packets regardless of its chipset and firmware.

\section{ML-Based CSI Estimation Method}
\label{sec:csi_recovery}
This section proposes the ML-based CSI estimation method with regard to BFMs.
Our technique to improve the estimation accuracy is to leverage an inter-subcarrier dependency, i.e., CSI and BFM at subcarrier $k$, $\bm{H}[k]$, and $\bm{V}[k]$, which are dependent on BFMs at neighboring subcarriers.
Thus, rather than the subcarrier-independent estimation, i.e., CSI $\bm{H}[k]$ estimation with respect to BFM $\bm{V}[k]$,
we utilize the subcarrier-integrated estimation, i.e., CSI $\Hseq$ estimation from  BFMs at all the subcarriers, $\Vseq$.
Therefore, we train the ML model so that it outputs CSI from BFMs at multiple subcarriers.
Moreover, to extract inter-subcarrier dependency, we have used ``frequency-directional convolutional layer'' as an input layer of the ML model.
The frequency-directional convolutional layer outputs the convolution result of the BFMs and a kernel filter along the direction of frequency.

\label{sec:learning_procedure}
\begin{figure}[t]
	\centering
	\includegraphics[width=1.0\columnwidth]{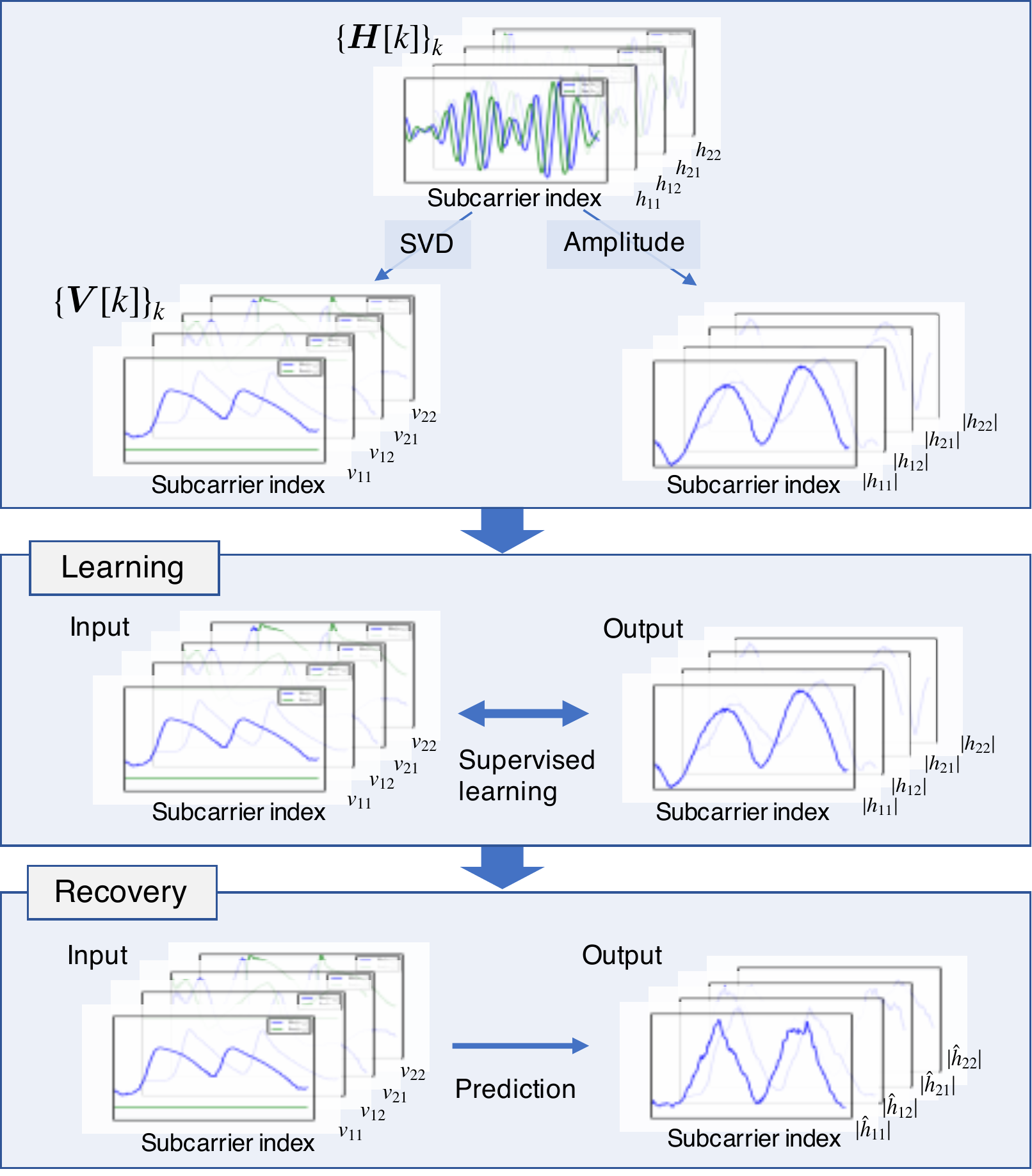}
	\caption{
		Procedure of CSI estimation.
		First, the amplitude of the CSI and BFMs $\Vseq$ are computed.
		Second, in the learning step, the model is trained using $\Vseq$ as inputs and amplitude of CSI as outputs.
		Finally, using the trained model, the amplitude of the original CSI is predicted based on the corresponding beamforming matrices $\Vseq$.
	}
	\label{fig:abst_of_recovery}
\end{figure}
Fig.~\ref{fig:abst_of_recovery} shows an overview of the recovery process.
Based on the CSI, BFMs are calculated using SVD; that is,
the dataset comprising BFMs and the corresponding CSI is generated.
Note that we have used the BFMs of all the subcarriers for the input feature.
Then, the ML model is trained using BFMs ($\Vseq$) as the input and the corresponding CSI ($\Hseq$) as the target label.

In this study, as a firsthand approach, the amplitude of the CSI elements ($|h_{ij}|$) is estimated.
The amplitude of the channel indicates attenuation in the channel between the transmit and receive antennas.
Indeed, several studies on CSI sensing studies \cite{wang2017carm, li2016csi, chen2018passive} have been conducted using the amplitude of CSI.
Hence, we believe that the feasibility of the amplitude estimation fills the gap between CSI and BFM-based sensing techniques.


\section{Performance Evaluation}
\label{sec:ml_algorithms}

\subsection{Channel Simulation}
\label{sec:simulation}
\begin{table}[t]
	\caption{Simulation Specifications}
	\centering
	\begin{tabular}{lr} \toprule
		Parameters                                   & Values    \\
		\midrule
		Number of transmit antennas $N_{\mathrm{t}}$ & 2         \\
		Number of receive antennas $N_{\mathrm{r}}$  & 2         \\
		Carrier frequency                            & 5.25\,GHz \\
		Delay profile                                & Model-B   \\
		Channel bandwidth $W$                        & 80\,MHz   \\
		\bottomrule
	\end{tabular}
	\label{tab:parameter}
\end{table}
The simulation parameters are listed in Table~\ref{tab:parameter}.
A transmitter with two antennas communicates with a receiver with two antennas using $2\times 2$ MIMO.
The channel model is TGac channel model \cite{TGac_path}.
We generate 10,000 different 
CSI matrices. 
The BFM matrix, which is calculated by following the procedure of 802.11ac\footnote{
	In 802.11ac/ax protocol, the BFM packet contains the quantized BFM data.
	However, as a firsthand study, this study ignores the effect of the quantization of the BFMs and assumes that the full-percept BFM can be obtained.}, is labeled by its corresponding CSI matrix.

The multiple CSI-BFM pairs are gathered as a data sample.
Considering that the CSI and BFMs have the same shape ($2\times 2$) and $N$ pairs are gathered,
the input and output data samples are tensors, whose shape is $N\times 2\times 2$.
The generated data samples are divided into 90\% and 10\%, and they are used to train and evaluate the ML model, respectively.

\subsection{Structure and Hyperparameters of the ML Algorithms}
\begin{figure}[t]
	\centering
	\includegraphics[width=6cm,height=7cm,clip]{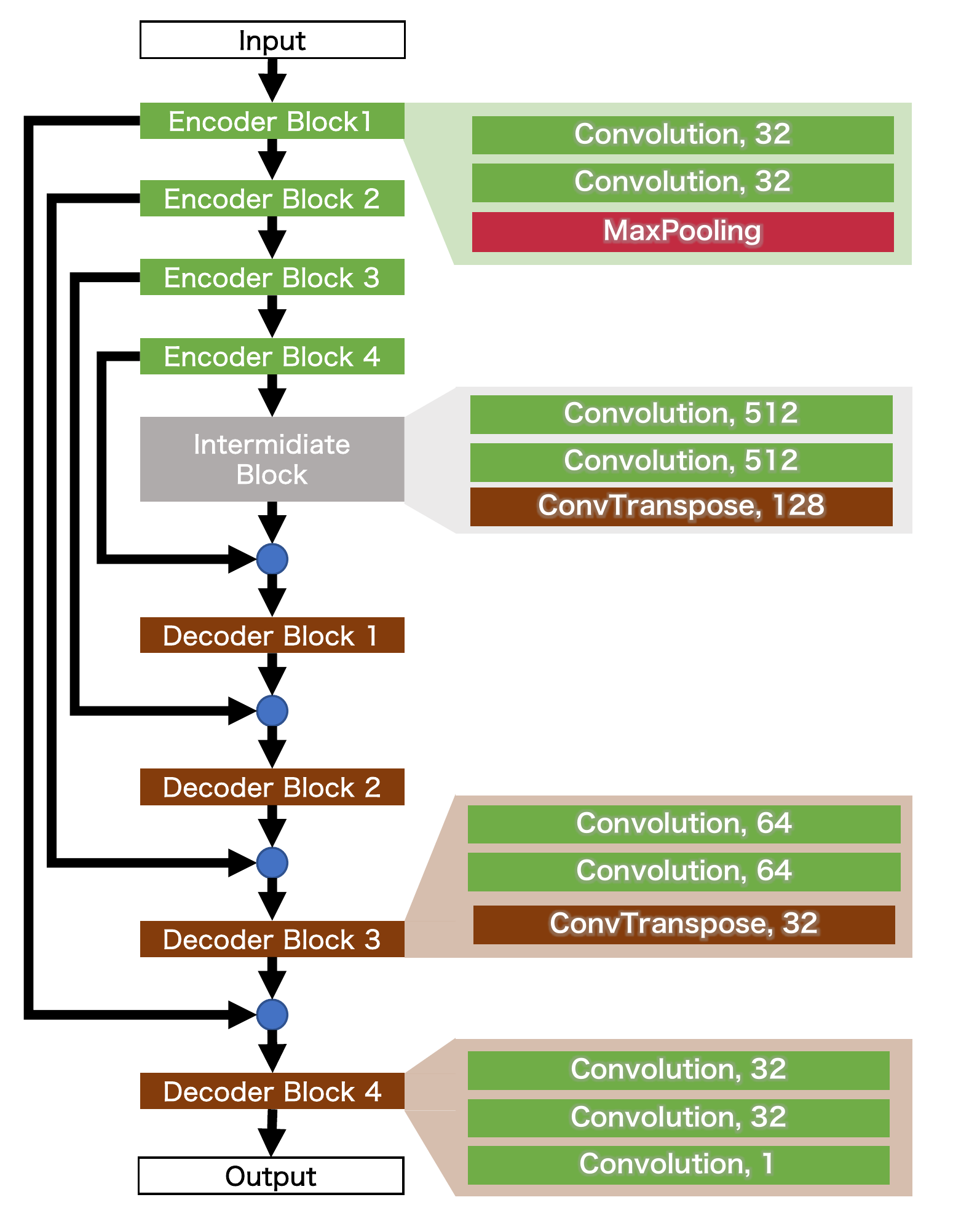}
	\label{fig:LearningModel}
	\caption{
		Overview of the CNN structure.
		The blue circle indicates channel-wise concatenation.
	}
	\label{fig:NNstructure}
\end{figure}

We examine two ML models for CSI estimation, namely, convolutional LSTM network (ConvLSTM) and convolutional network (CNN),
which are commonly used in the field of computer vision.
Fig.~\ref{fig:NNstructure} shows the structure of the CNN model.
The CNN model follows \cite{ronneberger:2015:UNET} and comprises four encoder blocks, an intermediate block, and four decoder blocks.
Each encoder block contracts the input to low-dimensional representation and comprises two convolution layers followed by the max pooling layer.
The intermediate block processes the representation and feeds its output to the decoder blocks.
For regularization, dropout and batch normalization layers follow the third and fourth encoder blocks.
Each decoder block expands the low-dimensional representation to the output shape and comprises two convolution layers and a convolutional transpose layer.
The convolutional transpose layer is a backward step in the normal convolution procedure.
The filter sizes of convolution layers, convolutional transpose layers, and max pooling layers are $6\times 2$, $6\times 2$, and $1\times 2$, respectively.
Except for the fourth encoder block, the output of the $i$th encoder block is fed to the $(5-i)$th decoder block and $(i+1)$th encoder block.
The output of the fourth encoder block is fed to the first decoder block and the intermediate block.
The number of output channels of the $i$th encoder block is the same as that of the $(5-i)$th decoder block (32, 64, 128, and 256 for first, second, third, and fourth encoder block, respectively).
The connections of the blocks and detail parameters are shown in Fig.~\ref{fig:NNstructure}.
Except for the intermediate block, the structure of ConvLSTM is the same as that of the CNN model.
The intermediate block of the CNN model is replaced by ConvLSTM layers with eight units.

As noted previously, the training dataset was divided into training and validation datasets in the ratio of 9:1.
The ML model is updated for multiple epochs using the training data only (and not the validation data).
In each epoch, the model is evaluated using the validation dataset.
The training is completed if 200 epochs are performed, or if the validation loss is increased after 10 epochs; it indicates that the model is starting to overfit.
When updating and distilling the models, mini-batch size, the number of epochs in each round, and the learning rate are selected as 64, 0.001, respectively.


\subsection{Results}

\begin{figure}[t]
	\centering
	\includegraphics[width=0.4\textwidth]{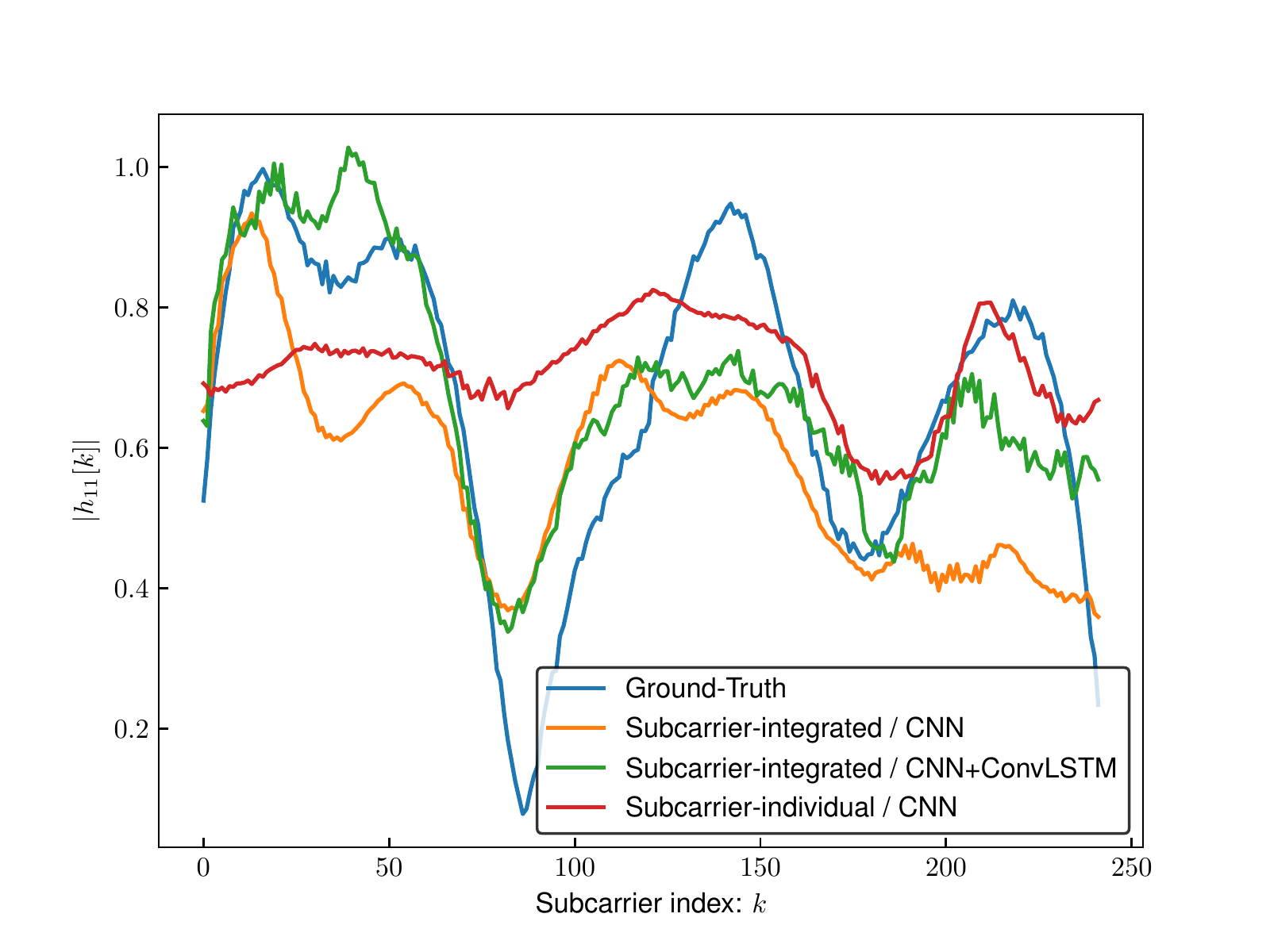}
	\caption{
	Ground truth (simulated amplitude of $\Hseq$) and recovered amplitude from frequency-series matrices $\Vseq$.
	Recovery of $|h_{11}|$ from frequency-series matrices $\Vseq$.
	}
	\label{fig:recovery_result}
\end{figure}

\begin{table}[t]
	\caption{Frobenius Error of Each CSI Estimation}
	\centering
	\begin{tabular}{ccc} \toprule
		Input shapes                     & ML models    & Frobenius errors \\
		\midrule
		Subcarrier-integrated estimation & CNN+ConvLSTM & 0.434            \\
		Subcarrier-integrated estimation & CNN          & 0.448            \\
		Subcarrier-individual estimation & CNN          & 0.539            \\
		\bottomrule
	\end{tabular}
	\label{tab:Error_esults}
\end{table}
\begin{figure}[t]
	\centering
	\includegraphics[width=0.35\textwidth]{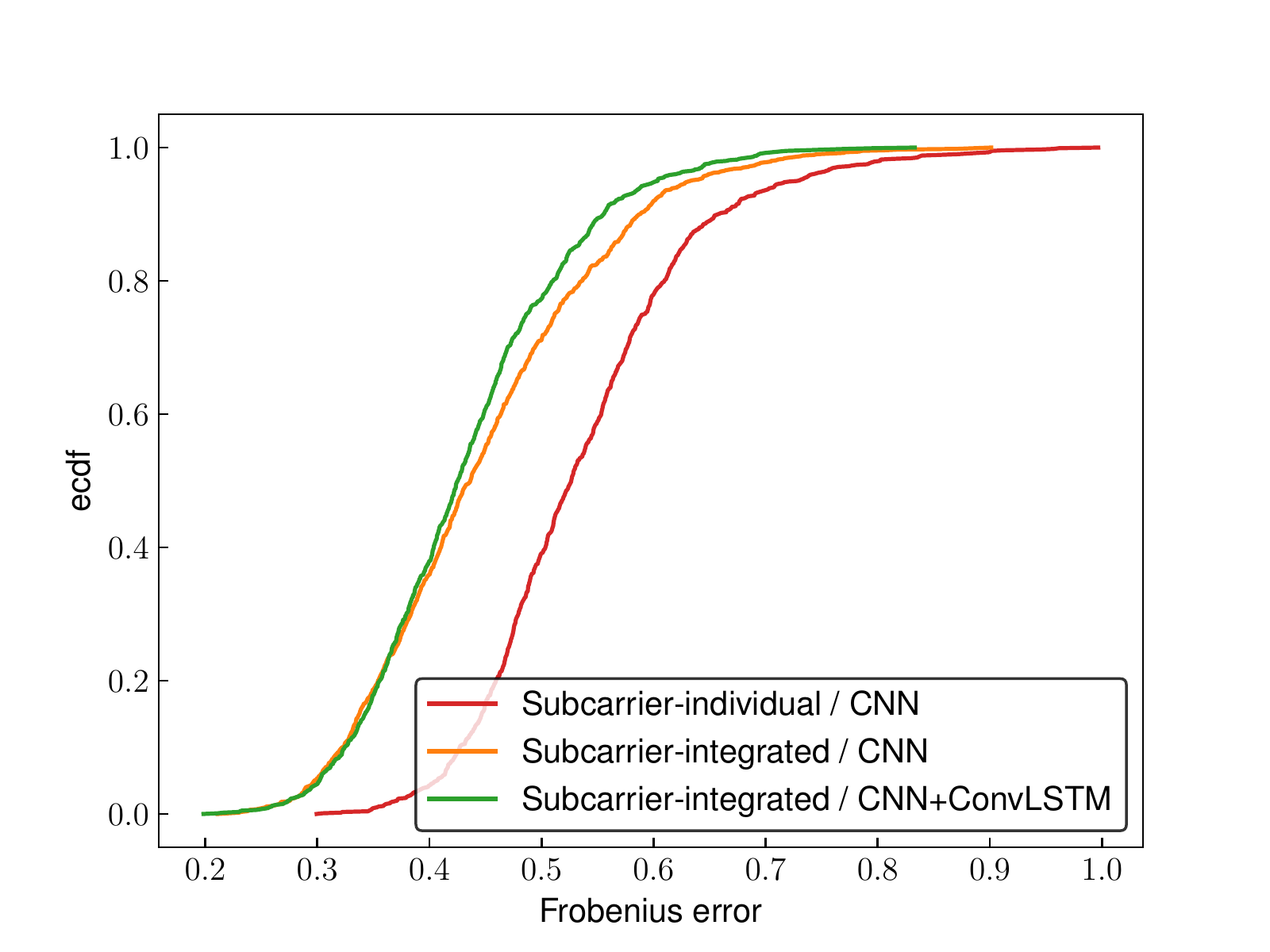}
	\caption{
		Cdf of Frobenius error of each CSI estimation.
	}
	\label{fig:cdf_frobenius}
\end{figure}
\begin{figure}[t]
	\centering
	\includegraphics[width=0.35\textwidth]{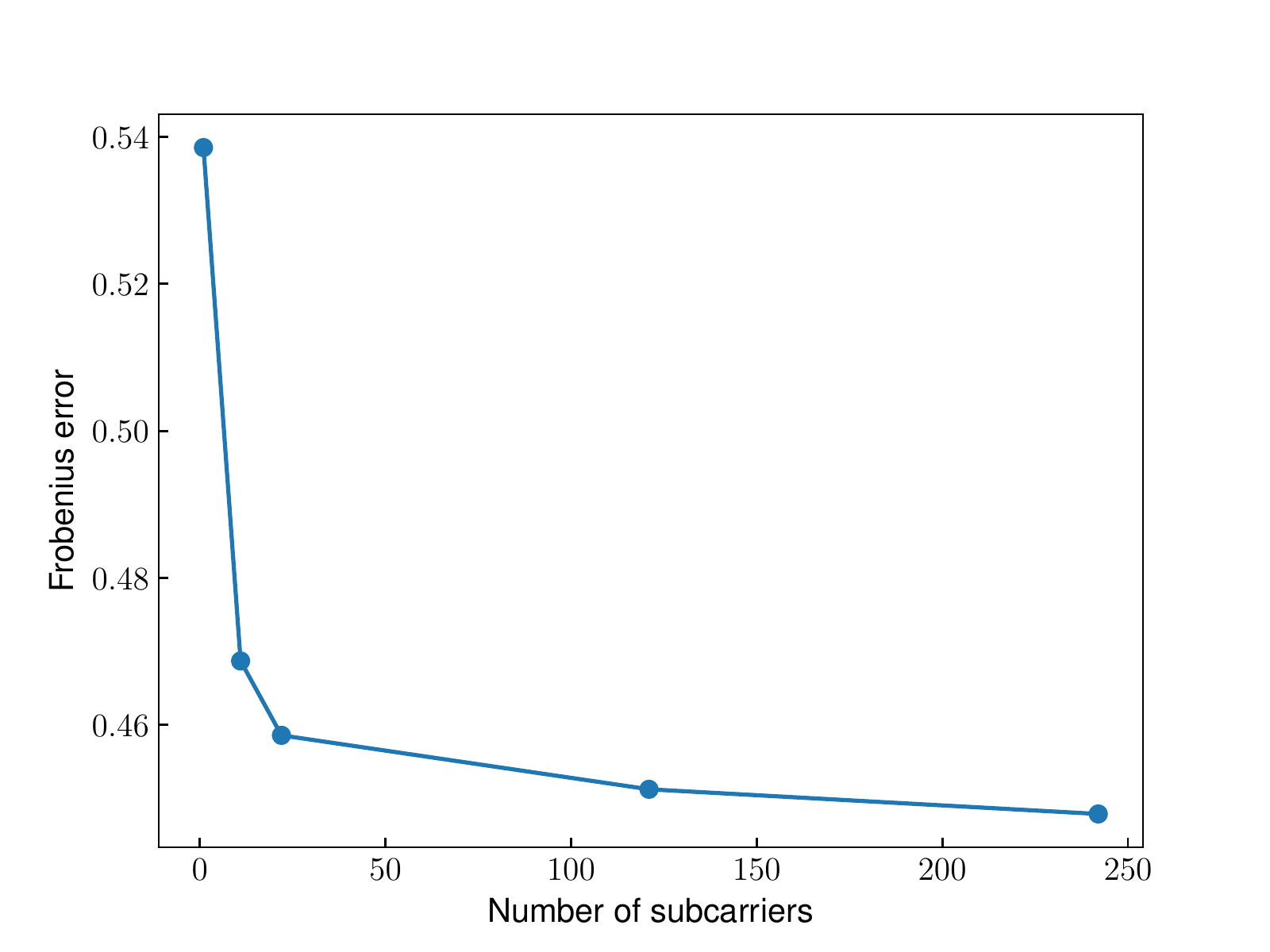}
	\caption{
		Frobenius error of each subcarrier.
		The scenario where the number of subcarriers is 1 corresponds to the subcarrier-individual estimation.
	}
	\label{fig:num_subcarrier}
\end{figure}

Fig.~\ref{fig:recovery_result} shows the frequency series of the ground-truth amplitude of CSI (i.e., CSI obtained from channel simulation) and that of the recovered CSI from the BFM based on each ML algorithm.
To examine our key idea to estimate CSI from BFMs of all the subcarriers 
rather than the subcarrier-independent estimation,
the estimation results of the subcarrier-integrated estimation are obtained by feeding all the BFMs at once to estimate the CSI of all the subcarriers, (i.e., the input and output data shape of the multi-subcarrier is $242 \times 2 \times 2$ tensor).
Conversely, that of subcarrier-individual estimation is obtained by feeding a single subcarrier's BFM  at once to estimate a single subcarrier's CSI,
(i.e., the input and output data shape of the single subcarrier is  $2 \times 2$ matrix).
The subcarrier-integrated estimation implies that, in the estimation process, the model uses the inter-subcarrier information, while the subcarrier-individual estimation uses 1 by 1 corresponding only.

The estimated frequency-selective amplitude matches the ground-truth amplitude of CSI,
while the estimated one based on the correspondence of the subcarrier-individual estimation incurs larger losses.

Table~\ref{tab:Error_esults} lists the average values of the Frobenius errors of each CSI estimation,
and Fig.~\ref{fig:cdf_frobenius} shows the empirical cumulative distribution function (ecdf) of Frobenius errors for test datasets.
The subcarrier-integrated estimation exhibits fewer errors than the subcarrier-individual estimation.
Moreover, being consistent with the aforementioned description, the CNN+ConvLSTM model achieved the smallest Frobenius error compared with the other models.

Fig.~\ref{fig:num_subcarrier} shows the Frobenius errors as a function of the number of input and output subcarriers.
The use of several subcarriers simultaneously has been suggested to improve the accuracy of CSI estimation.
In this estimation process, we have equated the data that are used to train the model.
Based on these results, leveraging the inter-subcarrier dependency of the BFM of the multi-subcarrier, the accuracy of CSI estimation with regard to BFMs has improved.

\section{Conclusion}
\label{sec:conclusion}
We evaluated the possibility of CSI recovery from BFMs specified in IEEE 802.11ac/ax.
The key idea to enable accurate estimation is to leverage the frequency-sequential BFMs
and to estimate CSI of BFMs at all the subcarriers, rather than estimating CSI of BFMs at a single subcarrier.
The simulation revealed that the estimated frequency-selective amplitude of CSI matches the ground-truth amplitude.
Our future studies include the evaluation of the proposed CSI estimation method in a real environment.

\section*{Acknowledgment}
This research and development work was partially supported by the MIC/SCOPE \#JP196000002 and JSPS KAKENHI (under Grant No. JP18H01442).

\bibliographystyle{IEEEtran}
\bibliography{ccnc_csi}

\end{document}